\documentclass[letterpaper, 10 pt, conference]{ieeeconf}  

\IEEEoverridecommandlockouts                              

\let\proof\relax \let\endproof\relax

\usepackage{amsthm}

\usepackage{times}
\usepackage{soul}
\usepackage{url}
\usepackage[hidelinks]{hyperref}
\usepackage[utf8]{inputenc}
\usepackage[small]{caption}
\usepackage{graphicx}

\usepackage{booktabs}
\usepackage{amsmath,amssymb,amsfonts}

\usepackage[ruled,vlined]{algorithm2e}
\usepackage{algorithmic}

\usepackage{graphicx}
\usepackage{textcomp}
\usepackage{xcolor}

\usepackage{subfig}
\usepackage{multirow}
\usepackage{makecell}
\usepackage{dblfloatfix}

\newtheorem{thm}{Theorem}

\theoremstyle{definition}
\newtheorem{defn}{Definition}

\usepackage{footnote}
\usepackage{makecell}

\title{\LARGE \bf
Hierarchical Constrained Stochastic Shortest Path Planning via Cost Budget Allocation
}

\author{Sungkweon Hong$^{1}$ and Brian C. Williams$^{1}$
\thanks{$^{1}$Computer Science and Artificial Intelligence Lab, Massachusetts Institute of Technology, Cambridge, MA 01239, USA, {\tt\small \{sk5050, williams\}@csail.mit.edu}. The work was supported by the Boeing Corporation.}
}

\begin{document}

\maketitle
\thispagestyle{empty}
\pagestyle{empty}

\begin{abstract}
Stochastic sequential decision making often requires hierarchical structure in the problem where each high-level action should be further planned with primitive states and actions. In addition, many real-world applications require a plan that satisfies constraints on the secondary costs such as risk measure or fuel consumption. In this paper, we propose a hierarchical constrained stochastic shortest path problem (HC-SSP) that meets those two crucial requirements in a single framework. Although HC-SSP provides a useful framework to model such planning requirements in many real-world applications, the resulting problem has high complexity and makes it difficult to find an optimal solution fast which prevents user from applying it to real-time and risk-sensitive applications. To address this problem, we present an algorithm that iteratively allocates cost budget to lower level planning problems based on branch-and-bound scheme to find a feasible solution fast and incrementally update the incumbent solution. We demonstrate the proposed algorithm in an evacuation scenario and prove the advantage over a state-of-the-art mathematical programming based approach.
\end{abstract}

\section{Introduction}\label{sec_intro}
As human beings, making decisions is one of the daily tasks to improve our lives, where many, if not most, decisions are sequential. Sequential decision making has been one of the core topics in Artificial Intelligence (AI) to support those pervasive tasks \cite{ernst2004power,silver2017mastering,cassandra1998survey}. The major role of sequential decision making is generating a course of actions that achieves the goals of an agent. Actions are defined from the domain of the problem and describe the states where they can be executed and how the states are changed by executing them. In commercial aircraft operation, for example, nominal unit operations are based on air route segments and standard departure/arrival procedures, and hence those can be reasonable choices of actions sets. 

In sequential decision making problems, a crucial assumption is that the actions are well defined and an agent is capable of executing those actions. For many real-world problems, however, such assumption is too strong to hold, especially for online applications. Again, in aircraft operations, even though each air route segment is well established and can be flown as a routine procedure, there might be a novel circumstance that prohibits an aircraft from flying the segment as published, such as convective weather conditions. In such situation, travelling the air route segment forms another planning problem, which makes the problem hierarchical. 

In fact, hierarchical structure in the planning problem has been widely studied in stochastic sequential decision making problems \cite{sutton1999between,theocharous2004approximate,lim2011monte,omidshafiei2017decentralized,amato2014planning}. In those works, options, also known as macro actions or skills, were introduced to abstract the underlying problem space into a higher-level space to scale both learning and planning processes. However, most of the works in hierarchical planning focus on minimizing the expected cost (or maximizing the expected reward). In many of the real-world applications, however, naively minimizing the expected cost might not be desirable. For example, when we want an aircraft to navigate to the destination as fast as possible, it is also important to ensure the aircraft can complete the mission with the onboard fuel or stays outside of the unsafe region for safety. 

Despite being non-hierarchical, stochastic sequential decision making problems have been widely extended to include constraints. One of the most well-known frameworks is constrained stochastic shortest path problems (C-SSPs) \cite{altman1999constrained}, which optimize the primary cost function while constraining the secondary cost functions. In this paper, we propose a new framework, hierarchical constrained stochastic shortest path problem (HC-SSP), that extends C-SSP to hierarchical problem structure. To the authors' knowledge, this extension has been little attempted, yet there is some work on this direction \cite{feyzabadi2015hcmdp}. However, the previous work 
focuses on abstracting the existing state space to find an approximated solution to the original problem. In our approach, on the other hand, we focus on finding (near-)optimal solution given the hierarchical structure of the problem. 

Although the extension from C-SSP to HC-SSP looks straightforward, this extension poses a unique challenge that has not been present in the previous works. A hierarchical approach has its advantage based on decoupled high-level actions that can be learned individually. However, with a constraint, they are no longer decoupled due to shared cost budget given by the constraint. To address this problem, we propose a cost budget allocation approach based on the well-known branch-and-bound scheme. 

This paper makes two main contributions. First, we propose a hierarchical constrained stochastic shortest path problem (HC-SSP) that extends C-SSP with hierarchical structure. Second, we propose a budget-allocation-based algorithm that finds a deterministic hierarchical policy to a HC-SSP in an anytime fashion. The algorithm is shown to be theoretically optimal given a specific condition and can be approximated based on the user-defined approximation level. The experimental results show that the algorithm can find near-optimal solutions in most cases even with high approximation level. 

The remainder of this paper is organized as follows. Section~\ref{sec_prem} summarizes the most relevant backgrounds to our work. Section~\ref{sec_hc_ssp} presents the proposed HC-SSP problem definition with an intuitive example. Section~\ref{sec_app} elaborates our proposed algorithm, followed by evaluation results in Section~\ref{sec_experiments}. 
Finally, we conclude and present future work in Section~\ref{sec_conclusion}.

\section{Preliminaries}\label{sec_prem}

\subsection{Constrained Stochastic Shortest Path Problem}

A \textit{constrained stochastic shortest path problem} (C-SSP) is a tuple $\langle \mathcal{S}, \bar{s}, \mathcal{G}, \mathcal{A}, T, \vec{C}, \vec{\Delta} \rangle$ in which $\mathcal{S}$ is a set of discrete states; $\bar{s}\in \mathcal{S}$ is the initial state; $\mathcal{G}\subset \mathcal{S}$ is a set of goal states; $\mathcal{A}$ is a set of discrete actions; $T : \mathcal{S} \times \mathcal{A} \times \mathcal{S} \rightarrow [0, 1]$ is the state transition function, where $T(s,a,s')=Pr(s'|s,a)$ is the probability of being in state $s'$ after executing action $a$ in state $s$; $\vec{C}$ is the indexed set of cost functions $\{C_0, ..., C_N\}$, where each $C_i:\mathcal{S} \times \mathcal{A} \times \mathcal{S} \rightarrow \mathbb{R}_+$ is the cost function and $C_i(s,a,s')$ is the $i$-th cost of executing action $a$ in the state $s$ and resulting in the state $s'$; and $\vec{\Delta}$ is the indexed set of bounds $\{\Delta_1,\dots,\Delta_N\}$, where $\Delta_i$ is the bound for $i$-th cost function for $i=1,\dots,N$ \cite{altman1999constrained,trevizan2016heuristic}. 

A solution to a C-SSP is a policy $\pi:\mathcal{S}\times \mathcal{A}\rightarrow [0,1]$ which maps states to a probability distribution over actions. If a policy maps every state to a probability distribution with a single outcome, the policy is called ``\textit{deterministic},'' and is called ``\textit{stochastic}'' otherwise. In the former case, we denote $\pi(s)$ as the action selected by the policy $\pi$ for the state $s$.

An optimal solution to a C-SSP is a policy $\pi^*$ which minimizes the total expected cost, but in addition, the expected cost of the secondary cost function $C_i$ should be upper bounded by $\Delta_i$ for $i=1,\dots,N$. In other words, an optimal policy $\pi^*$ for a C-SSP is defined as follows:
\begin{align*}
	\underset{\pi\in \Pi}{\text{argmin}}\qquad & f(\pi) \\
	\text{s.t.}\qquad & g_i(\pi) \leq 0 \quad \text{for $i=1,\dots,N$},
\end{align*}
where $\Pi$ is the set of all policies and 
\begin{align*}
    f(\pi) &= \mathbb{E} \Bigg[ \sum_{t=0}^{\infty} C_0(s_t,a_t,s_{t+1}) \bigg| s_0=\bar{s}, \pi \Bigg]  \\
    g_i(\pi) &= \mathbb{E} \Bigg[ \sum_{t=0}^{\infty} C_i(s_t,a_t,s_{t+1}) \bigg| s_0=\bar{s}, \pi\Bigg] - \Delta_i.
\end{align*}

Note that every optimal policy for a C-SSP might be stochastic \cite{altman1999constrained}. In this paper, however, we limit the policy space to deterministic ones which are preferred due to explainability and reliability in the safety-critical applications such as aircraft operations. Note that it is known that limiting the policy space to deterministic ones significantly increases the complexity of the problem \cite{feinberg2000constrained}.

\subsection{Anytime Algorithm for C-SSPs}\label{sec_anytime_alg}
In this section, we introduce a recently developed anytime algorithm for C-SSPs in \cite{hong2021anytime} which we use as a subroutine in our proposed method. The algorithm has two stages where the first stage finds a Lagrangian dual solution of a C-SSP and then the second stage incrementally closes a duality gap, if there is any. In the first stage, the Lagrangian function is defined as follows:
\begin{align}
    L(\lambda, \pi) = f(\pi) + \lambda \cdot g(\pi),
    \label{eq:lag_1}
\end{align}
where $\lambda=[\lambda_1,\dots,\lambda_N]\in\mathbb{R}_+^N$ and $g=[g_1,\dots,g_N]^\top$. Then the Lagrangian dual problem is defined as follows:
\begin{align}
    L^* = L(\lambda^*) = \max_{\lambda \geq 0} L(\lambda),
    \label{eq:lag_4}
\end{align}
where
\begin{align}
    L(\lambda) = \min_{\pi\in\Pi} L(\lambda, \pi).
    \label{eq:lagrangian_function}
\end{align}
Note that the following statement holds, according to the \textit{weak duality} \cite{boyd2004convex}:
\begin{align}
    L(\lambda) \leq f^{*}, \quad \forall \lambda \geq 0,
\end{align}
where $f^{*}$ is the primal optimal cost. If $L(\lambda) = f^{*}$, where the \textit{strong duality} holds, the dual solution coincides with a true optimal solution. However, strong duality usually does not hold in our problem and a duality gap exists. The purpose of the second stage is to reduce such gap and find a primal optimal solution. 

\begin{figure}[t!]
  \centering
  \includegraphics[width=0.8\linewidth]{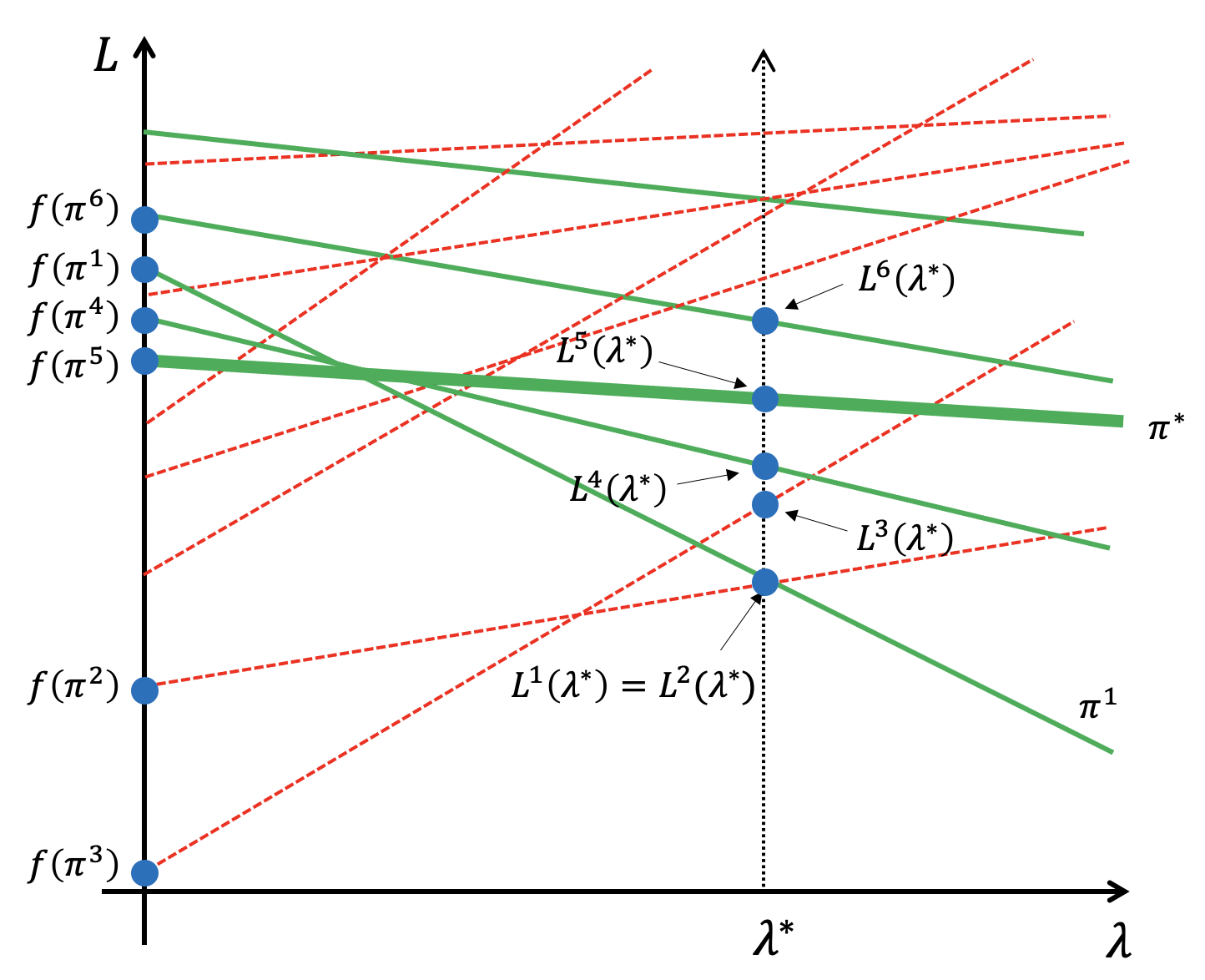}
  \caption{Visualization of the second stage of the anytime algorithm.}
  \label{fig:closing-gap}
\end{figure}

Conceptually, the way the algorithm obtains a primal optimal policy is quite simple. From the dual optimal policy, we can find a primal optimal solution by iteratively finding the next best solutions, with respect to the Lagrangian function value $L(\lambda^*, \pi)$. This procedure is illustrated in Fig.~\ref{fig:closing-gap} with an intuitive example which only includes a single secondary cost function. The figure shows all the deterministic policies in $(\lambda, L)$ space, including feasible ones ($g_1(\pi)\leq 0$) and infeasible ones ($g_1(\pi)>0$), in green solid lines and red dotted lines, respectively. Note that the true primal optimal policy $\pi^*$ is indicated with a bold line. Fig.~\ref{fig:closing-gap} also shows the Lagrangian function value $L(\lambda^*,\pi)$ for each policy as an intersection of the policy with the dotted vertical line at $\lambda^*$. Similarly, a $y$-intercept of a policy $\pi$ is the primal cost $f(\pi)$. In addition, $\pi^1$ is the solution associated with $\lambda^*$ which can be obtained from the first stage, where $\pi^k$ denotes $k$-th best policy with respect to the Lagrangian function value $L(\lambda^*, \pi)$. Starting from $\pi^1$, we find the next best policies in terms of $L(\lambda^*, \pi)$, in a non-decreasing manner, until we obtain a primal optimal policy $\pi^*$, and update the incumbent policy whenever we obtain a better feasible policy. Note that the Lagrangian function value and incumbent cost at any time set lower and upper bounds on the optimal cost of the C-SSP, respectively. We refer to \cite{hong2021anytime} for the details of the anytime algorithm.

\section{Hierarchical Constrained Stochastic Shortest Path Problem}\label{sec_hc_ssp}
In this section, we define a \textit{hierarchical constrained stochastic shortest path problem} (HC-SSP). We begin this section with an intuitive example of a HC-SSP followed by a formal definition. 

\subsection{HC-SSP in a Nutshell}\label{sec_nutshell}

\begin{figure}
  \centering
  \includegraphics[width=0.8\linewidth]{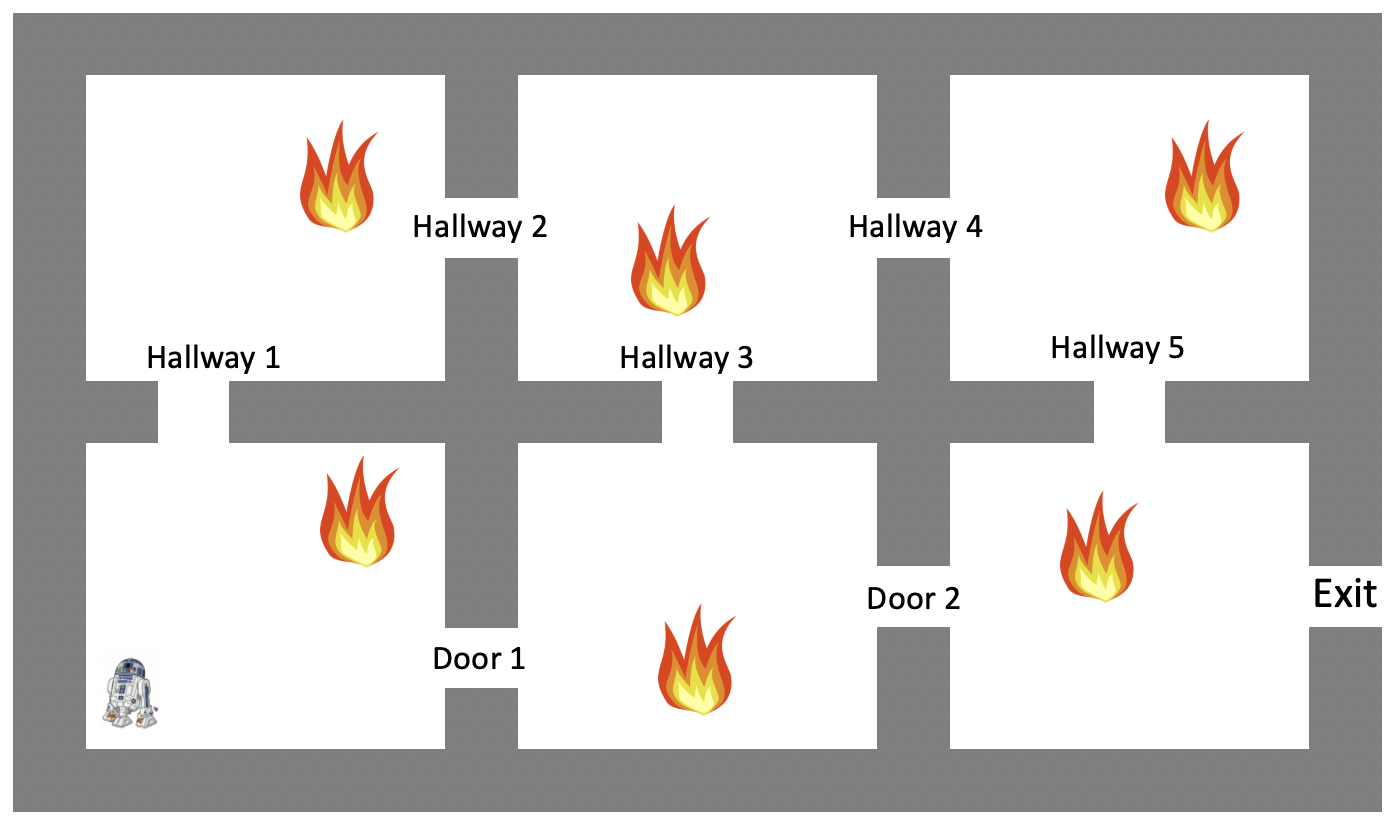}
  \caption{Illustration of the robot evacuation scenario.}
  \label{fig:evacuation}
\end{figure}

\begin{figure}
  \centering
  \includegraphics[width=0.9\linewidth]{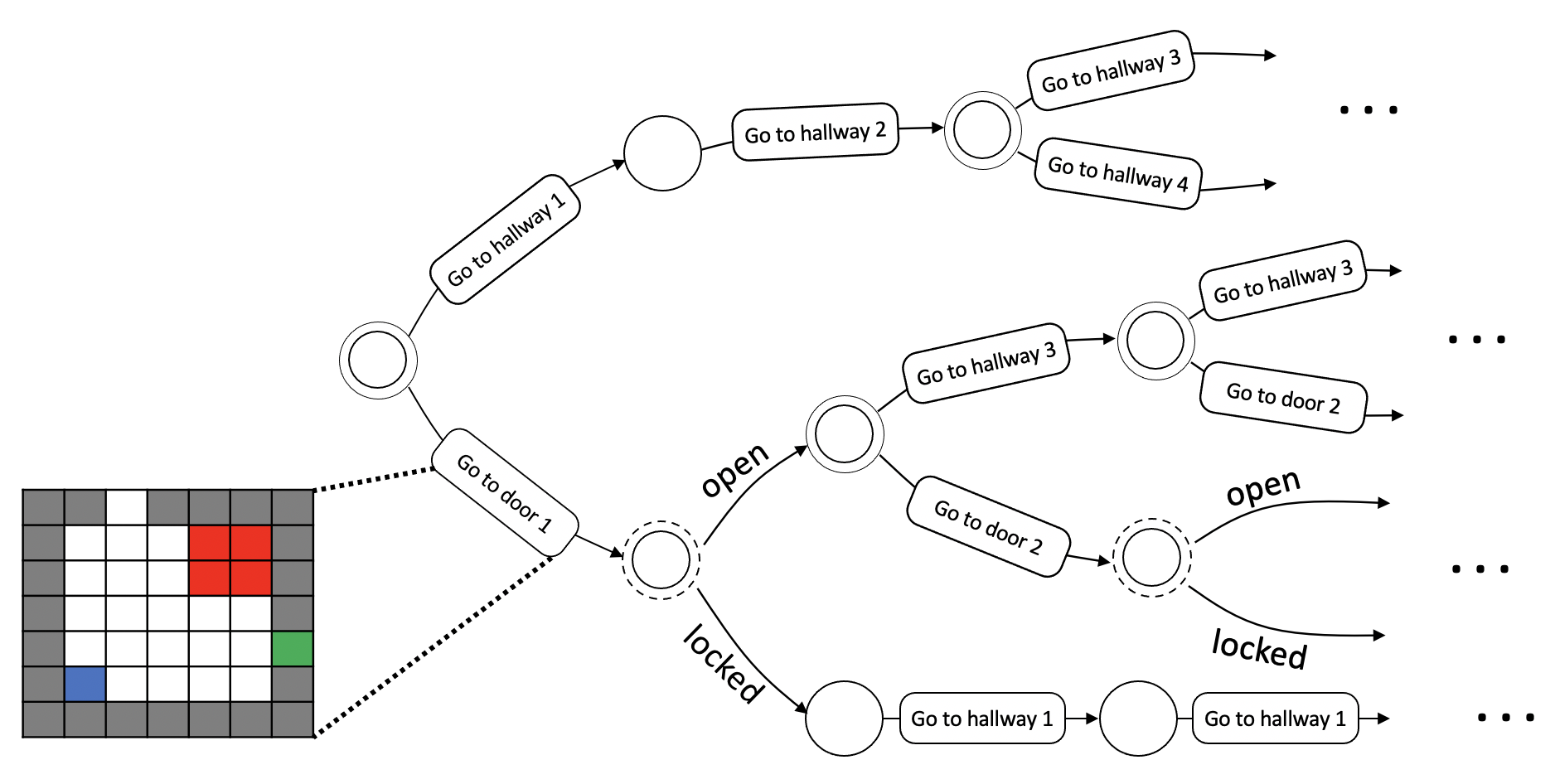}
  \caption{Graphical representation of the procedural planning problem for the evacuation scenario in Fig.~\ref{fig:evacuation}. An example of planning problem for \texttt{go-to-door1} activity is also shown, where red shaded states are hazardous states and blue and green states are initial and goal states, respectively.}
  \label{fig:procedural_planning}
\end{figure}

Consider a grounded example shown in Fig.~\ref{fig:evacuation}, in which a robot is evacuating from a building with six rooms in a hazardous environment. In this example, a robot should make a series of decisions about which hallway or door to approach to get to the exit point. At a quick glance, using door 1 and door 2 seems to let the robot evacuate faster than the other options. However, the decision should be made carefully since the robot has only partial information on the building: the robot knows that the hallways are always clear, but also knows that the doors are locked with a 0.1 probability, in which case the robot should take a roundabout way. By assuming that a set of activities that move a robot from a specific landmark to another is well defined, the problem can be formulated as a conditional procedural planning problem, which is shown in Fig.~\ref{fig:procedural_planning} graphically. Note that, each circle and oval represents a notable event and activities, respectively. Also note that, double circle and dotted double circle represent controllable and uncontrollable choices, respectively.

One of the keys to our approach is noting that each activity (e.g., \texttt{go-to-door1}) is not necessarily predefined. Rather, each of the activities can be defined as an another planning problem with its own state and action space. For example, each activity can be considered as a stochastic shortest path problem in a grid world as shown in Fig.~\ref{fig:procedural_planning}.

Another key to our approach is constraints on secondary costs defined over a set of activities. Consider a global constraint for the evacuation scenario is defined verbally as follows: ``\textit{The damage caused by fire must be less than or equal to $\Delta$ until evacuation.}'' With such constraint, the planning problem for \texttt{go-to-door1} in Fig.~\ref{fig:procedural_planning} is now a constrained stochastic shortest path problem.

There is another yet more important implication of having such global constraint. Due to the global constraint, activities are no longer independent. Suppose we are planning for an activity \texttt{go-to-door1} as a sub-problem for our evacuation planning problem. Then, how much of the cost out of the global damage budget $\Delta$ should be allocated for this activity? For example, using budget parsimoniously in planning \texttt{go-to-door1} might result in a conservative and sub-optimal plan if optimal solutions for other activities cause negligible damage. On the other hand, using too much of the budget in planning \texttt{go-to-door1} might result in sub-optimality as well, since other activities could reduce global cost by using more budget. This motivates us to allocate the budget systematically for different activities to obtain global optimality which will be discussed in more detail in Section~\ref{sec_app}.

Finally, a solution to HC-SSP is a hierarchical policy that consists of both procedural policy and policies for every associated activity. There are two desirable properties for a solution. First, the solution should be feasible, which means that the solution should not violate any constraint. In the evacuation scenario, we only have one global constraint that requires the damage caused by fire to be less than or equal to $\Delta$. Therefore, the weighted sum of the damage cost for every activity that is selected by a procedural policy should be less than or equal to $\Delta$, where a weight for an activity is the likelihood of executing that activity. Second, the solution should minimize the expected cost. It is not always possible, however, to obtain an optimal solution, especially when the planning time is limited. Therefore, we want to develop a solution method with an anytime property which can output an incumbent feasible solution at timeout while the solution eventually converges to a true optimal solution with sufficient planning time. 

\subsection{Problem Definition}\label{sec_prob_def}

A HC-SSP is a tuple $\langle \mathcal{S}, \mathcal{I}, \mathcal{T}, \mathcal{V}, \mathcal{E}, \mathcal{P}, \{\mathcal{C}_j\}_{j\in J} \rangle$, where
\begin{itemize}
    \item $\mathcal{S}$ is a set of discrete states.
    
    \item $\mathcal{I}:\mathcal{S}\rightarrow [0,1]$ is the initial distribution.
    
    \item $\mathcal{T}$ is a set of notable events. Note that $t^s$ and $t^e$ are two special events, where $t^s$ is the start event and $t^e$ is the end event. 
    
    \item $\mathcal{V}$ is a set of choice variables, each $v\in \mathcal{V}$ being a discrete, finite domain variable. Note that each $v\in \mathcal{V}$ is associated with an event $t\in\mathcal{T}$, where we denote $\mathcal{V}(t)=v$.
    
    \item $\mathcal{E}$ is a set of activities. An activity $E\in \mathcal{E}$ is activated if both start and end events of the activity are activated. 
    
    \item $\mathcal{P}:\mathcal{T}\times \bigcup_{t\in\mathcal{T}} \text{Dom}\{ \mathcal{V}(t) \} \times \mathcal{T}\rightarrow [0,1]$ is a probabilistic transition function between events based on the assignments of choice variables, where $\mathcal{P}(t,a,t')=Pr(t'|t,a)$ is the probability of being in event $t'$ after executing choice $a$ in event $t$.
    
    \item $\{\mathcal{C}_j\}_{j\in J}$ is an indexed set of constraints, where $J=\{1,\dots,M\}$ and $M$ is the number of constraints. Each $\mathcal{C}_j$ is a tuple $\langle \Psi_j, I_j, \Delta_j  \rangle$, where $\Psi_j\subset \mathcal{E}$ is a set of activities associated with the constraint $\mathcal{C}_j$, $\Delta_j$ is an upper bound and $I_j:\Psi_j\rightarrow \mathbb{N}$ is an index function for each $E\in \Psi_j$ that indicates which secondary cost function is used for $\mathcal{C}_j$. Note that in this paper we assume that a secondary cost function of an activity is included in at most one constraint for simplicity. This assumption, however, can be easily generalized with little modification. 
    
\end{itemize}

A HC-SSP abstracts the underlying state-action space and dynamics based on events and activities, where the activities are selected by choice variables and the transitions between events are governed by transition function $\mathcal{P}$. Then each activity is defined over a subset of underlying state space and has its own action space and dynamics with its sub-goal. Formally, an activity $E\in \mathcal{E}$ is a tuple $\langle t_E^s, t_E^e, \mathcal{S}_E, \mathcal{A}_E, T_E, \vec{C}_E, \mathcal{G}_E \rangle$, where
\begin{itemize}
    \item $t_E^s, t_E^e\in \mathcal{T}$ are the start and end events of the activity, respectively.

    \item $\mathcal{S}_E\subset \mathcal{S}$ is a set of discrete states.
    
    \item $\mathcal{A}_E$ is a set of discrete actions.
    
    \item $T_E:\mathcal{S}_E\times \mathcal{A}_E\times \mathcal{S}_E\rightarrow [0,1]$ is the state transition function, where $T_E(s,a,s')=Pr(s'|s,a)$ is the probability of being in state $s'$ after executing action $a$ in state $s$.

    \item $\vec{C}_E=\{C_E^0, ..., C_E^{N_E}\}$ is an indexed set of cost functions, where each $C_E^i:\mathcal{S}_E\times \mathcal{A}_E\times \mathcal{S}_E\rightarrow \mathbb{R}_+$ is the $i$-th cost function that defines the $i$-th cost of executing action $a$ in state $s$ and resulting in state $s'$ and $N_E$ is the number of secondary cost functions for the activity $E$. 
    
    
    \item $\mathcal{G}_E\subset \mathcal{S}_E$ is the set of goal state. 
    
\end{itemize}

In this paper, we refer to the planning problem over the events and activities as \textit{procedural planning} and the planning problem for each activity as \textit{activity planning}, which will be discussed in more detail in Section~\ref{sec_reduction}.

\subsection{Solution to HC-SSP}
\label{sub:output}
A solution to a HC-SSP problem is a tuple $\langle \rho, \Gamma \rangle$, where
\begin{itemize}
    \item $\rho$ is a procedural policy which fully assigns every choice variable $\mathcal{V}$.
    
    \item $\Gamma$ is a set of policies for every activated activity by $\rho$, where a policy $\Gamma_E:\mathcal{S}_E\rightarrow \mathcal{A}_E$ is a mapping from states to actions for an activity $E$. 
\end{itemize}

Note that, abusing the notation, $E\in\rho$ if an activity $E\in\mathcal{E}$ is activated by $\rho$, and $E\not\in\rho$ otherwise. 
Then, an optimal solution to a HC-SSP is defined as follows:
\begin{align}
    \underset{\rho, \Gamma}{\text{argmin}} \quad &\mathbb{E}_{E\sim \rho} \Big[ f^E\big( \Gamma_E \big) \Big] \\
    \text{s.t.}\quad & \mathcal{I} = \Gamma_{E}^I \text{ for } E\in\rho \text{ such that } t_{E}^s = t^s, \label{eq:flow_1} \\
    \quad & \Gamma_E^T = \Gamma_{E'}^I \text{ for } E,E'\in\rho \text{ such that } t_E^e = t_{E'}^s, \label{eq:flow_2} \\
    \quad & \sum_{E\in \Psi_j} L(E|\rho)\cdot g_{I_j(E)}^E\big( \Gamma_E \big)  \leq \Delta_j \text{ for } j\in J,
\end{align}
where $\Gamma_E^I$ and $\Gamma_E^T$ are initial and termination distributions of the policy $\Gamma_E$ for an activity $E$, respectively, $L(E|\rho)$ is a likelihood of executing an activity $E$ given procedural policy $\rho$, and $f^E( \Gamma_E )$ is the expected primary cost for an activity $E$ which is defined as follows:
\begin{align*}
    f^E( \Gamma_E ) = \mathbb{E} \Bigg[ \sum_{t=0}^{\infty} C_E^0(s_t,a_t,s_{t+1}) \bigg|\Gamma_E^I, \Gamma_E \Bigg],
\end{align*}
and $g_i^E( \Gamma_E )$ is the expected $i$-th secondary cost for an activity $E$ which is defined as follows:
\begin{align*}
    g_i^E( \Gamma_E ) = \mathbb{E} \Bigg[ \sum_{t=0}^{\infty} C_E^i(s_t,a_t,s_{t+1}) \bigg|\Gamma_E^I, \Gamma_E \Bigg].
\end{align*}

\section{Approach}\label{sec_app}
In this section, we present our proposed approach for a HC-SSP. We begin with an intuitive introduction of the approach and then present algorithms and formal analysis. 

\subsection{Overview}
\begin{figure}[h!] 
    \centering
    \subfloat[Example procedural planning problem with two activities.]{%
        \includegraphics[width=0.23\textwidth]{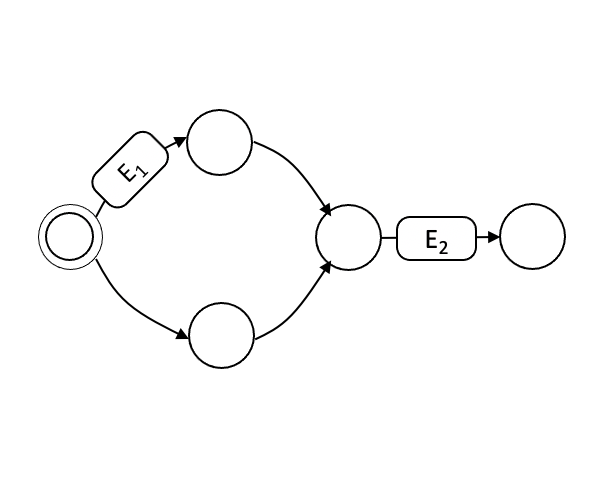}%
        \label{fig:alloc_ex_1}%
        }%
    \hfill%
    \subfloat[Budget allocation space where $x-$ and $y-$ axes are allocated cost budget for $E_1$ and $E_2$.]{%
        \includegraphics[width=0.23\textwidth]{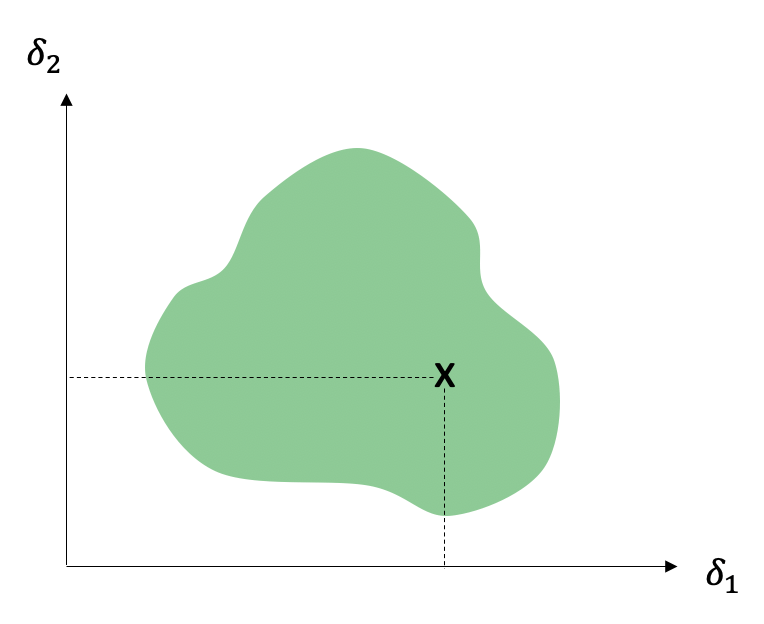}%
        \label{fig:alloc_ex_2}%
        }%
    \caption{Illustration of budget allocation for an example with only two activities and a single global constraint.}
    \label{fig:alloc_ex}
\end{figure}

For the purpose of illustration, let us consider a simple problem shown in Fig.~\ref{fig:alloc_ex_1} which has two activities and a single constraint including both activities. Our approach is based on the fact that we can decouple the procedural and each activity planning problems if the constraint budget $\Delta$ is allocated properly to each of the activities.

Fig.~\ref{fig:alloc_ex_2} shows the constraint budget allocation space for the example problem, where $x-$ and $y-$axes are the allocated budget for activities $E_1$ and $E_2$, respectively, and the green shaded region conceptually shows the region where a feasible solution to HC-SSP exists. Then if we allocate a specific budget to $E_1$ and $E_2$, respectively, then we can find policies for each activity based on the allocation, which in turn allows us to find procedural policy based on $f$ and $g$ values of each activity. However, since the allocation space is continuous, it is computationally intractable to check every possible budget allocation to find an optimal solution. Instead, we frame the allocation problem based on the branch-and-bound scheme \cite{balakrishnan1991branch,horst2013global}.

Let $M^-_{E, I_j(E)}$ and $M^+_{E, I_j(E)}$ be lower and upper limits of budget allocation for a constraint $\mathcal{C}_j$ and an activity $E\in\Psi_j$. Then the allocation space that we are interested in, which we denote as $\mathcal{Q}_\text{init}$, is defined as follows:
\begin{align*}
    \mathcal{Q}_\text{init} = \prod_{j\in J} \prod_{E\in\Psi_j} \Big[ M^-_{E, I_j(E)}, M^+_{E, I_j(E)} \Big].
\end{align*}
Note that a simple choice of such lower and upper limits for $E\in\Psi_j$ is $0$ and $\frac{1}{L(E)}\cdot \Delta_j$, where $L(E)$ is the minimum likelihood of activity $E$, although a more sophisticated bound can be found based on pre-processing. Then the branch-and-bound algorithm keeps splitting $\mathcal{Q}_\text{init}$ into smaller partitions and computes upper ($\alpha$) and lower bounds ($\beta$) for the newly generated partitions, which can be used not only to guide the search toward an optimal solution but also to determine the quality of the current incumbent solution. 

Algorithm~\ref{algo:branch_and_bound} shows the branch-and-bound algorithm for a HC-SSP, where $\alpha(\mathcal{Q})$ and $\beta(\mathcal{Q})$ denote upper and lower bounds for a partition $\mathcal{Q}$, respectively, and $\alpha_k$ and $\beta_k$ denotes global upper and lower bounds at the $k$-th iteration, respectively. After initialization (lines 1--9), the algorithm splits a partition with the lowest lower bound into two partitions along its longest edge to form a new partition (lines 11--13). Then it computes lower and upper bounds for newly generated partitions (lines 14--19) and updates the global lower and upper bounds (lines 20--21). In addition, whenever a better feasible solution is found, it updates the incumbent solution (lines 4--6 and 16--18). 

The core parts of Algorithm~\ref{algo:branch_and_bound} are Compute-LB and Compute-UB subroutines which highly affect the performance of the algorithm. Before presenting the proposed Compute-LB and Compute-UB subroutines, let us begin with introducing procedural and activity planning in more detail, which has been briefly introduced in Section~\ref{sec_prob_def}.

\begin{algorithm}[t!]
\DontPrintSemicolon
\LinesNumbered 
\KwIn{HC-SSP instance $\mathcal{H}$, initial budget allocation space $\mathcal{Q}_\text{init}$, approximation parameter $l$, optimality tolerance $\epsilon$}
  $k=0$ \;
  $\mathcal{M}_0 = \{ \mathcal{Q}_{\text{init}} \}$ \;
  $\alpha(\mathcal{Q}_{\text{init}}), \langle \rho_0, \Pi_0 \rangle \leftarrow$ Compute-UB($\mathcal{H}$, $\mathcal{Q}_{\text{init}}$, $l$) \;
  \If{$\langle \rho_0, \Pi_0 \rangle$ is not $\textbf{None}$}{
  $\langle \rho_\text{inc}, \Pi_\text{inc} \rangle \leftarrow \langle \rho_0, \Pi_0 \rangle$ \;
  $\alpha_\text{inc} \leftarrow \alpha(\mathcal{Q}_\text{init})$
  }
  $\beta(\mathcal{Q}_{\text{init}}) \leftarrow$ Compute-LB($\mathcal{H}$, $\mathcal{Q}_{\text{init}}$, $l$) \;
  $\alpha_0 \leftarrow \alpha(\mathcal{Q}_{\text{init}})$ \;
  $\beta_0 \leftarrow \beta(\mathcal{Q}_{\text{init}})$ \;
  \While{$\alpha_k - \beta_k > \epsilon$}{
    $\mathcal{Q}_k \leftarrow \text{argmin}_{\mathcal{Q}\in\mathcal{M}_k} \beta(\mathcal{Q})$ \;
    split $\mathcal{Q}_k$ along one of its longest edges into $\mathcal{Q}'_k$ and $\mathcal{Q}''_k$ \;
    $\mathcal{M}_{k+1} \leftarrow \big(\mathcal{M}_k \backslash \{\mathcal{Q}_k\}\big) \cup \big(\mathcal{Q}'_k\cup \mathcal{Q}''_k\big)$ \;
    \For{$\mathcal{Q}\in \{\mathcal{Q}'_k, \mathcal{Q}''_k\}$}{
      $\alpha(\mathcal{Q}), \langle \rho', \Pi' \rangle \leftarrow$ Compute-UB($\mathcal{H}$, $\mathcal{Q}$, $l$) \;
      \If{$\alpha(\mathcal{Q}) < \alpha_\text{inc}$}{
        $\langle \rho_\text{inc}, \Pi_\text{inc} \rangle \leftarrow \langle \rho', \Pi' \rangle$ \;
        $\alpha_\text{inc} \leftarrow \alpha(\mathcal{Q})$
      }
      $\beta(\mathcal{Q}) \leftarrow$ Compute-LB($\mathcal{H}$, $\mathcal{Q}$, $l$)}
    $\alpha_k \leftarrow \min_{\mathcal{Q}\in\mathcal{M}_k}\alpha(\mathcal{Q})$ \;
    $\beta_k \leftarrow \min_{\mathcal{Q}\in\mathcal{M}_k}\beta(\mathcal{Q})$ \;
    $k\leftarrow k+1$ \;
  }
  \Return $\langle \rho_\text{inc}, \Pi_\text{inc} \rangle$
 \caption{Branch-and-Bound}
 \label{algo:branch_and_bound}
\end{algorithm}

\subsection{Procedural and Activity Planning}\label{sec_reduction}
In this section, we define \textit{procedural planning} and \textit{activity planning}, and then show that both of the planning problems can be reduced to an instance of C-SSP. Let us begin with the definition of procedural planning.

\defn{(Procedural Planning)}
Suppose estimates for the primary and $i$-th secondary cost functions for every activity $E\in \mathcal{E}$ and $i\in \{1,\dots,N_E\}$ are given, where the former and the latter are denoted as $\bar{f}^E$ and $\bar{g}_i^E$, respectively. Then we define planning for a procedural plan $\rho$ based on the given estimates as \textit{procedural planning}.

Similarly, we define activity planning as follows.

\defn{(Activity Planning)}
For $E\in \mathcal{E}$, suppose the initial distribution, $\delta_i^-$ and $\delta_i^+$ are given for $i=1,\dots,N_E$, where $\delta_i^-$ and $\delta_i^+$ are lower and upper bounds on $i$-th cost function. Then we define planning for an activity $E$ based on the given bounds as \textit{activity planning}.

Now, we show that both procedural and activity planning are reducible to C-SSPs in the following theorem.

\begin{thm}
Both procedural and activity planning can be reduced to an instance of C-SSP. \end{thm}

\noindent\textit{Proof Sketch.}
First, it is obvious that an activity planning is an instance of a C-SSP given initial distribution and bounds on the secondary cost functions. 

Now, given $\bar{f}^E$ and $\bar{g}_i^E$ for every $E\in\mathcal{E}$ and $i\in \{1,\dots,N_E\}$, it can be shown that a procedural planning can be reduced to a C-SSP by matching each element as follows: $\mathcal{S} = \mathcal{T}$, $\bar{s} = \mathcal{I}$, $\mathcal{G} = \{t^e\}$. $\mathcal{A} = \bigcup_{t\in\mathcal{T}} \text{Dom}\{\mathcal{V}(t)\}$, $T(t,a,t') = \mathcal{P}(t,a,t')$, $\vec{C} = [C_0, C_1, \dots, C_J]$, where
\begin{equation*}
    C_0(t,a,t') = \begin{cases} 
    \bar{f}^E & \parbox[t]{1.6in}{if $\exists E\in\mathcal{E}$ such that $t_E^s=t$ and $t_E^e=t'$, } \\
    0 & \text{otherwise},
    \end{cases}
\end{equation*}
and
\begin{align*}
    C_j(t,a,t') = \begin{cases} 
    \bar{g}_i^E & \parbox[t]{1.6in}{if $\exists E\in\mathcal{E}$ such that $t_E^s=t$, $t_E^e=t'$ and $E\in \Psi_j$, } \\
    0 & \text{otherwise},
    \end{cases}
\end{align*}
for every $j\in J$, and $\vec{\Delta}=[\Delta_1,\dots,\Delta_J]$. \qed

Note that we denote $\bar{f}$ as a collection of $\bar{f}^E$ for every $E\in \mathcal{E}$, and similarly, $\bar{g}$ as a collection of $\bar{g}_i^E$ for every $E\in \mathcal{E}$ and $i\in \{1,\dots,N_E\}$. 

Finally, we provide algorithms for activity and procedural planning in Algorithm~\ref{algo:activity-planning} and \ref{algo:procedural-planning}. Both algorithms basically instantiate a C-SSP and solve it based on the anytime algorithm to produce lower and upper bounds along with the incumbent policy, as introduced in Section~\ref{sec_anytime_alg}. The parameter $l$ used in the algorithms governs the approximation level of the anytime algorithm by limiting the number of iterations for the second stage of the algorithm. For example, if $l=0$, then the algorithm returns a Lagrangian dual solution, and if $l=\infty$, then the algorithm runs until it converges and returns a true optimal solution. 

\begin{algorithm}[h!]
\DontPrintSemicolon
\LinesNumbered 
\KwIn{Activity $E$, partition $\mathcal{Q}$, approximation parameter $l$}
  Instantiate a C-SSP based on $E$ and $\mathcal{Q}$ \\
  Run stage 1 of the anytime algorithm to produce $LB$, $UB$ and $\pi_\text{inc}$ \\
  \If{$l>0$}{
  Run stage 2 of the anytime algorithm for $l$-iterations to update $LB$, $UB$ and $\pi_\text{inc}$
  }
  \Return $\pi_\text{inc}$, $LB$, $UB$
 \caption{Activity-Planning}
 \label{algo:activity-planning}
\end{algorithm}

\begin{algorithm}[h!]
\DontPrintSemicolon
\LinesNumbered 
\KwIn{HC-SSP problem instance $\mathcal{H}$, primary cost estimate $\bar{f}$, secondary cost estimate $\bar{g}$, approximation parameter $l$}
  Instantiate a C-SSP based on $\mathcal{H}$, $\bar{f}$ and $\bar{g}$ \\
  Run stage 1 of the anytime algorithm to produce $LB$, $UB$ and $\pi_\text{inc}$ \\
  \If{$l>0$}{
  Run stage 2 of the anytime algorithm for $l$-iterations to update $LB$, $UB$ and $\pi_\text{inc}$
  }
  \Return $\pi_\text{inc}$, $LB$, $UB$
 \caption{Procedural-Planning}
 \label{algo:procedural-planning}
\end{algorithm}

\subsection{Computing Lower and Upper Bounds}
In this section, we provide methods for computing lower and upper bounds on the optimum of a HC-SSP given a partition $\mathcal{Q}$, which are used as subroutines in Algorithm~\ref{algo:branch_and_bound}. 

Algorithm~\ref{algo:compute-lb} shows Compute-LB subroutine, where $\mathcal{Q}_{E,i}^-$ denotes a lower limit of the $i$-th secondary cost for an activity $E$ defined by a partition $\mathcal{Q}\subset\mathcal{Q}_{\text{init}}$. Given a partition $\mathcal{Q}$ and an approximation parameter $l$, Algorithm~\ref{algo:compute-lb} performs the activity planning for each activity in a topological order so that the flow constraints in Eqs.~\eqref{eq:flow_1} and \eqref{eq:flow_2} are satisfied. To obtain a lower bound, the primary cost of each activity is estimated as a lower bound ($\phi_E^-$) computed from Algorithm~\ref{algo:activity-planning}. In addition, the secondary cost of each activity is estimated as a lower limit of the given partition. 

Similarly, Algorithm~\ref{algo:compute-ub} shows Compute-UB subroutine. Different from Compute-LB, Algorithm~\ref{algo:compute-ub} estimates the primary cost of each activity based on an upper bound ($\phi_E^+$) computed from Algorithm~\ref{algo:activity-planning}, which is the associated expected primary cost of the incumbent solution. In addition, the secondary cost is estimated as the expected secondary cost of the corresponding incumbent solution. Finally, Algorithm~\ref{algo:compute-ub} collects policies for each activity to form a solution to the HC-SSP. The following theorem formally shows that both Algorithm~\ref{algo:compute-lb} and \ref{algo:compute-ub} correctly compute lower and upper bounds. 

\begin{thm}\label{thm_lb_ub}
Algorithm~\ref{algo:compute-lb} computes a lower bound on the optimum to the given HC-SSP within the given partition $\mathcal{Q}$. Similarly, Algorithm~\ref{algo:compute-ub} computes an upper bound on the optimum to the given HC-SSP within the given partition $\mathcal{Q}$.
\end{thm}

\noindent\textit{Proof Sketch.}
First, Algorithm~\ref{algo:compute-lb} computes lower bounds on the optimal expected primary cost of each activity given the secondary cost bounds by $\mathcal{Q}$ and uses them as cost estimates (lines 2 and 3). In addition, the algorithm uses lower limits of the given partition $\mathcal{Q}$ as the secondary cost estimates (line 5). Let $\rho^*$ and $\Gamma_E^*$ be optimal procedural and activity policies in $\mathcal{Q}$. Let also $\Phi^-$ be the lower bound provided by Algorithm~\ref{algo:compute-lb}. Then
\begin{align*}
    \mathbb{E}_{E\sim \rho^*} \Big[ f^E\big( \Gamma_E^* \big) \Big] \geq \mathbb{E}_{E\sim \rho^*} \big[ \bar{f}^E \big] \geq \Phi^-,
\end{align*}
where the first inequality is based on the fact that $\bar{f}$ and $\bar{g}$ underestimate the cost functions, and the second inequality comes from the fact that the anytime algorithm used for a procedural planning gives a lower bound with any $l$. 

For Algorithm~\ref{algo:compute-ub}, it is easy to see that an upper bound is based on a feasible solution if one has been found, or infinity otherwise, which is obviously an upper bound on the optimum within the partition $\mathcal{Q}$. \qed

\begin{algorithm}[h!]
\DontPrintSemicolon
\LinesNumbered 
\KwIn{HC-SSP problem instance $\mathcal{H}$, partition $\mathcal{Q}$, approximation parameter $l$}
  \For{$E\in\mathcal{E}$ in a topological order}
  {$\Gamma_E, \phi_E^-, \phi_E^+ \leftarrow$ Activity-Planning($E, \mathcal{Q}$, $l$)\\
  $\bar{f}^E\leftarrow \phi_E^-$
  }
  \For{$j\in J$ and $E\in\Psi_j$}
  {$\bar{g}_{I_j}^E\leftarrow \mathcal{Q}_{E,I_j(E)}^-$}
  $\rho, \Phi^-, \Phi^+ \leftarrow$ Procedural-Planning($\mathcal{H}, \bar{f}, \bar{g}$, $l$) \;
  \Return $\Phi^-$
 \caption{Compute-LB}
 \label{algo:compute-lb}
\end{algorithm}

\begin{algorithm}[h!]
\DontPrintSemicolon
\LinesNumbered 
\KwIn{HC-SSP instance $\mathcal{H}$, partition $\mathcal{Q}$, approximation parameter $l$}
  $\Gamma = \{\}$ \;
  \For{$E\in\mathcal{E}$ in a topological order}
  {
  $\Gamma_E, \phi_E^-, \phi_E^+ \leftarrow$ Activity-Planning($E, \mathcal{Q}$, $l$)\\
  $\bar{f}^E\leftarrow \phi_E^+$\\
  $\Gamma\leftarrow \Gamma\cup \{\Gamma_E\}$
  }
  \For{$j\in J$ and $E\in\Psi_j$}
  {$\bar{g}_{I_j}^E\leftarrow g_{I_j}^E(\Gamma^E)$}
  $\rho, \Phi^-, \Phi^+ \leftarrow$ Procedural-Planning($\mathcal{H}, \bar{f}, \bar{g}$, $l$) \;
  \uIf{$\rho$ is None}{
  \Return $\infty, \text{None}$
  }
  \Else{
  \Return $\Phi^+$, $\langle \rho, \Gamma \rangle$
  }
 \caption{Compute-UB}
 \label{algo:compute-ub}
\end{algorithm}

\subsection{Convergence}
Now, we show that Algorithm~\ref{algo:branch_and_bound} is \textit{convergent}, i.e., for every $\epsilon>0$, $\alpha_k - \beta_k<\epsilon$ is satisfied within finite iterations given a certain condition. For this, we leverage a convergence analysis in \cite{balakrishnan1991branch} which showed that the branch-and-bound algorithm is convergent if the following two conditions are satisfied:
\begin{enumerate}
    \item[(R1)] $\beta(\mathcal{Q}) \leq f^*(\mathcal{Q}) \leq \alpha(\mathcal{Q})$. In other words, the functions Compute-LB and Compute-UB compute lower and
    upper bounds on $f^*(\mathcal{Q})$, respectively, where $f^*(\mathcal{Q})$ is the optimum within partition $\mathcal{Q}$. 
    
    \item[(R2)] Let $|\mathcal{Q}|$ be the maximum half-length of the sides of $\mathcal{Q}$. Then, as $|\mathcal{Q}|$ goes to zero, $\alpha_k - \beta_k$ uniformly converges to zero, i.e., for every $\epsilon>0$, there exist $\delta>0$ such that $\alpha(\mathcal{Q}) - \beta(\mathcal{Q})<\epsilon$ for every $\mathcal{Q}\subset \mathcal{Q}_\text{init}$ with $|\mathcal{Q}|\leq \delta$.
\end{enumerate}

Note that we already showed that (R1) is satisfied in Theorem~\ref{thm_lb_ub}. Now we conclude that Algorithm~\ref{algo:branch_and_bound} is convergent under a certain condition by the following theorem.
\begin{thm}
Suppose we use $l=\infty$ for Algorithm~\ref{algo:compute-lb}. Then Algorithm~\ref{algo:branch_and_bound} satisfies (R2).
\end{thm}

\noindent\textit{Proof Sketch.} First, recall that there is only a finite number of feasible solutions to a HC-SSP since we only consider deterministic policies. Let the budget allocation space $\mathcal{Q}_\text{init}$ has the dimension of $n$. Also, let $Y$ be a set of solutions to a HC-SSP, and $Z$ be a set of budget allocation points in $\mathcal{Q}_\text{init}$ correspond to $Y$. Then there exist $\epsilon>0$ such that $B(z,\epsilon)$ is disjoint for every $z\in Z$, where $B(x,r)$ is an open ball at $x$ with radius $r$, which implies that $||z_1 - z_2||_2 > 2\epsilon$ for every $z_1,z_2\in Z$. Then given the fact that $\sqrt{n}||x - y||_{\infty}\geq ||x - y||_2$, we know that $||z_1 - z_2||_{\infty} > \frac{2\epsilon}{\sqrt{n}}$ for every $z_1, z_2\in Z$.

Now, suppose $|\mathcal{Q}|\leq \frac{2\epsilon}{\sqrt{n}}$ for some $\mathcal{Q}\subset \mathcal{Q}_\text{init}$. Then at most a single point in $Z$ is in $\mathcal{Q}$. Suppose $z\in \mathcal{Q}\cap Z$ and $z$ corresponds to a feasible solution. Then the upper bound computed from Algorithm~\ref{algo:compute-ub} is the value of a feasible solution, which corresponds to $z$. Otherwise, i.e., if $z$ corresponds to an infeasible solution or $\mathcal{Q}\cap Z=\emptyset$, then the upper bound is $\infty$.

For each solution $y\in Y$, let $\eta_y>0$ be the maximum error in any estimate of $\bar{g}$ that does not affect the feasibility of the solution $y$. Then, let $\eta=\max_{y\in Y} \eta_y$. Now, let $\delta = \min\{\frac{2\epsilon}{\sqrt{n}}, \eta\}$ and $|\mathcal{Q}|\leq \delta$ for some $\mathcal{Q}\subset \mathcal{Q}_\text{init}$. Again, at most a single point in $Z$ is in $\mathcal{Q}$. Suppose $z\in \mathcal{Q}\cap Z$ and $z$ corresponds to a feasible solution. Since $l=\infty$, the lower bounds computed from activity plannings in Algorithm~\ref{algo:compute-lb} (line 3) are exact, and thus the lower bound is equal to the value of the corresponding feasible solution. Otherwise, i.e., if $z$ corresponds to an infeasible solution or $\mathcal{Q}\cap Z=\emptyset$, then the lower bound is $\infty$. Therefore, for any $\mathcal{Q}\subset \mathcal{Q}_\text{init}$ with $|\mathcal{Q}|\leq \delta$, $\alpha(\mathcal{Q}) - \beta(\mathcal{Q})=0$. \qed

Note that although $l=\infty$ is required for the optimality, running the anytime algorithm until its convergence for every iteration can be computationally intractable. Therefore, in practice, we can use small $l$ to balance the level of approximation. In Section~\ref{sec_experiments}, we show that even with $l=0$, the algorithm finds near-optimal solutions in most cases.

\section{Empirical Evaluation}\label{sec_experiments}

We begin this section with details of the evaluation scenario and then we provide the evaluation results to demonstrate the performance of the proposed method. 

\subsection{Evaluation Scenario}
We test our algorithm with the evacuation problem shown in Fig.~\ref{fig:evacuation}. In this problem, navigating between adjacent landmarks comprises a set of activities, where landmarks include initial position, exit position, doors and hallways. As mentioned in Section~\ref{sec_nutshell}, every hallway is known to be clear, but each door can be locked with a 0.1 probability.

Then, each room is modeled as a grid world and each activity is defined over a corresponding room. For each activity, there is one action for each cardinal direction, with a 0.85 probability of correct movement, and a 0.075 probability of +90 or -90 degrees deviation from the intended movement. The robot stays at the same state if it bumps into walls, which are boundaries of the grid world.

In addition, there are hazardous states which cause the 50 numerical damage to the robot. Then the primary objective is to evacuate as fast as possible while the constraint is to keep the damage level below the given threshold $\Delta$. More details of the scenario and the source code can be found at https://github.com/sk5050/HCSSP.

\subsection{Evaluation Results}

We evaluate the proposed method with the state-of-the-art MILP-based method \cite{dolgov2005stationary,ccpomdp_milp}. The benchmark MILP method first aggregates the rooms in Fig.~\ref{fig:evacuation}, which results in a single grid world with the additional \texttt{open} action next to the doors. Note that the total number of states in the aggregated problem is 18644. Then the evacuation problem is encoded as a MILP formulation and then solved using Gurobi 9. For more details about the encoding, please refer to \cite{dolgov2005stationary,ccpomdp_milp}. 

The proposed algorithm was implemented in Python 3, and all of the algorithms were executed on an Intel core i5-7200U with 8GB of RAM. In addition, the approximation parameter $l$ for the algorithm was set as 0 throughout the experiments. Finally, we used simple Euclidean distance and 0 for the primary and secondary cost heuristics, respectively, for the anytime algorithm. 

Table~\ref{tab:results} summarizes the evaluation results. We evaluated both MILP-based and the proposed method for different levels of $\Delta$'s, which is shown in the first column of the table. For each case, we show the results for both the proposed algorithm and MILP-based baseline method. For the baseline method, optimal objective and computation time in seconds are reported. The computation time includes both time for MILP encoding and the Gurobi solving time.

Similarly, for the proposed method, the optimality gap of the solution and the computation time are presented. To show the anytime history, three different solutions are reported, at the time when 10\%, 20\% and 50\% of the computation time has been spent compared to the computation time spent by the MILP-based method. We left the cell with a hyphen if no feasible solution has been found at the corresponding time bound. 

\begin{table}[h!]
  \scriptsize
  \centering
  \begin{tabular}{ |p{0.5cm}|p{0.5cm}|p{0.5cm}|p{0.5cm}|p{0.5cm}|p{0.5cm}|p{0.5cm}|p{0.5cm}|p{0.5cm}|}
    \hline
    \multirow{3}{*}{$\Delta$} & \multicolumn{2}{c|}{MILP}& \multicolumn{6}{c|}{Proposed method} \\
    \cline{2-9}
     & \multicolumn{2}{c|}{Opt}& \multicolumn{2}{c|}{10\%}& \multicolumn{2}{c|}{20\%}& \multicolumn{2}{c|}{50\%} \\
    \cline{2-9}
     & obj & time & gap & time & gap & time & gap & time \\
    \hline 
    0.01 & 145.71 & 404.95 & 0.00\% & 25.77 & 0.00\% & 25.77 & 0.00\% & 25.77 \\
    \hline 
    0.1 & 145.64 & 203.35 & - & - & 0.05\% & 24.27 & 0.05\% & 24.27 \\
    \hline 
    1 & 145.11 & 211.27 & - & - & 0.07\% & 39.20 & 0.07\% & 39.20 \\
    \hline 
    2 & 144.70 & 127.31 & - & - & 0.35\% & 21.95 & 0.35\% & 38.05 \\
    \hline 
    5 & 143.48 & 143.48 & - & - & 1.19\% & 21.85 & 1.19\% & 21.85 \\
    \hline 
    8 & 142.68 & 135.75 & - & - & 1.74\% & 21.56 & 0.45\% & 27.69 \\
    \hline 
    10 & 142.18 & 336.44 & 0.04\% & 21.88 & 0.04\% & 21.88 & 0.04\% & 21.88 \\
    \hline
  \end{tabular}
  \caption{Evaluation results of the evacuation scenario.}
  \label{tab:results}
\end{table}

As shown in the table, the proposed method could obtain a feasible solution within $20\%$ of the computation time compared to the MILP-based method. This result highlights the value of the proposed method in the practical domains where obtaining a safe policy is extremely important given highly limited planning time. Even further, it is interesting to note that the proposed method could obtain $<1\%$ of the optimality gap with less than half of the computation time compared to the MILP-based method except for one problem instance. In addition, the algorithm could obtain $<0.1\%$ of the optimality gap for 5 out of 7 cases with only up to $20\%$ of the baseline computation time. The worst case still only has $1.19\%$ of the optimality gap with $20\%$ of the baseline computation time. This result shows that the proposed algorithm is able to not only find a feasible solution faster than the non-hierarchical method, but also obtain near-optimal solutions even with small approximation parameter $l$ and simple heuristics.

\section{Conclusion}\label{sec_conclusion}

In this paper, we propose a hierarchical constrained stochastic shortest path problem (HC-SSP) that extends C-SSP to a hierarchical structure. Then we present a branch-and-bound based algorithm with a notion of cost budget allocation. Although the algorithm is proved to be optimal, the true value of the algorithm is its anytime property with high convergence rate to a near-optimal solution even with high approximation level. 

We demonstrate the advantage of the proposed algorithm compared to a MILP-based method in an evacuation scenario, yielding 2--5 times speed-ups with less than 1.2\% of the optimality gaps. Future work includes more comprehensive experiments in various problem domains. Another promising future work is allowing a nested hierarchical structure in the problem.

\bibliographystyle{IEEEtran}
\bibliography{IEEEabrv,mybib}

\end{document}